# PIC-Net: Point Cloud and Image Collaboration Network for Large-Scale Place Recognition

Yuheng Lu, Fan Yang, Fangping Chen, Don Xie
National Engineering Laboratory for Video Technology, Peking University
`{yuhenglu, fyang.eecs, chenfangping, donxie}@pku.edu.cn`

## Abstract

*Place recognition is one of the hot research fields in automation technology and is still an open issue, Camera and Lidar are two mainstream sensors used in this task, Camera-based methods are easily affected by illumination and season changes, LIDAR cannot get the rich data as the image could , In this paper, we propose the PIC-Net (Point cloud and Image Collaboration Network), which use attention mechanism to fuse the features of image and point cloud, and mine the complementary information between the two. Furthermore, in order to improve the recognition performance at night, we transform the night image into the daytime style. Comparison results show that the collaboration of image and point cloud outperform both image-based and point cloud-based method, the attention strategy and day-night-transform could further improve the performance.*

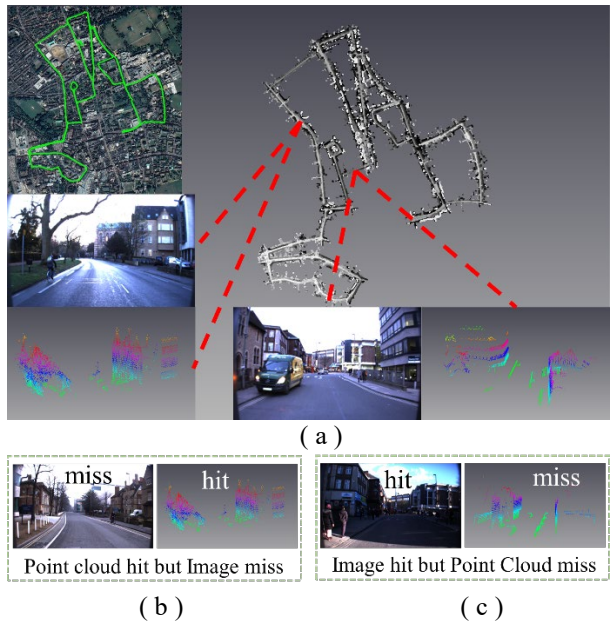

Figure 1. (a) The upper left of the image is the driving trajectory of Oxford RobotCar dataset [1], and the main image on the right is the final point cloud map. The image and point cloud pair in the lower left and the lower right are the two submaps. (b) top-1 retrieval result of lower left image and point cloud, respectively. (c) top-1 retrieval result of lower right image and point cloud, respectively.

## 1. Introduction

Given an image or point cloud, the place recognition task needs to locate where this data was collected. This task is still an open-issue in the field of Robotics, which can help to bridge the gap between assisted-driving and self-driving car, UAV, etc. Camera and Lidar are the two main data sources for place recognition task. Image-based place recognition methods are inherently affected by weather, light, season and other changes, and are easily to fail in these changes. Lidar can obtain the structure information of the environment, and the feature is more stable than image. However, point cloud is sparse and cannot capture complete structure of scene.

Similar to image retrieval, the place recognition task needs to build a database, and retrieve with query images. The database consists of a series of images, which are evenly distributed in the real world, each image in the database is represent with a one hot global features, thus place recognition task turns into how to extract discriminative global features. Global features are usually aggregated from local features, which could extract using handcrafted feature like SIFT, ORB, or learned deep feature. Bag of Words(BOW) [2][3], Vector of Locally Aggregated Descriptors (VLAD) [4] are two commonly used local feature aggregation methods, but they are not in end to end manner. NetVLAD [5] improves VLAD and transfers the local feature aggregation into a learnable manner.

Recently, Lidar technology has gradually matured. It has become an indispensable sensor in application scenarios, such as self-driving car, mapping, etc. Lidar point cloud is more robust to illumination change and its data is more reliable. Place recognition based on Lidar point cloud has attracted many researchers recently, but the point cloud in this task is usually sparse, and cannot capture the complete structure of surrounding. How to fill the gaps between points? Image may be a good option because of the high resolution. However, many factors would affect the image feature, such as dark night or overexposure. In this condition, image features will become the noise of global feature. Thus, the method to determine the reliability of those two kinds of data is needed.

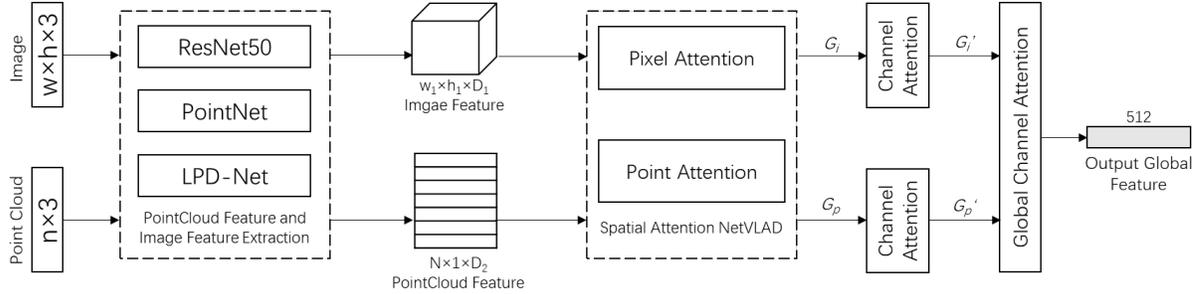

Figure 2. PIC-Net is composed of image and point cloud branch. The image branch uses Resnet50 to extract image feature map, and the point cloud branch uses PointNet [6] or LPD-Net [7] (removing NetVLAD [5] layer) to extract each point feature. Then, the image and point cloud features pass through the spatial attention layer, the channel attention layer and the global channel attention layer to obtain the final global feature.

In this paper, we propose the Point cloud and image collaboration network for Large-Scale Place Recognition, which use global channel attention to enhance the reliable feature inter image and point cloud feature, and use spatial attention VLAD to select the discriminative points and pixels. Our contributions include: 1) propose the point cloud and image collaboration network for the feature fusion and achieve state-of-art performance. 2) transform the night image into the daytime domain which further improve the performance. 3) prove that point cloud and image are complementary and can enhance each other.

## 1. RELATED WORK

In this section, we review other place recognition works, including image-based, point cloud-based and image and point cloud fusion-based place recognition methods.

### 2.1 Image based place recognition

Image based place recognition is similar to image retrieval, image classification, person re-identification tasks, the difference is mainly in the annotation of data, and the core is to extract the most discriminative features from given images. The typical feature extraction pipeline is to extract SIFT feature descriptors, and fuse these local SIFT features into global features, which include vectors of locally aggregated descriptors(VLAD) [4], bag-of-word(BOW) [2][3], Fish Vector[8] and Hamming embedding[9].

With the widespread application of Convolutional Neural Networks(CNN) in various image processing tasks, some researchers have used CNN feature to replace SIFT features in place recognition task. Arandjelovic et al.[5] propose NetVLAD, which uses the idea of VLAD feature extraction into end-to-end CNN based framework. First, it uses ResNet/VGG to extract the feature map with dimensions of W × H × D, which can be regarded as a feature of t × D dimension (t equals W × H ), next, clusters the t features into k clusters and calculates the residuals between features and cluster centers, then normalizes the residuals of each cluster center, finally, cascades the normalized residuals into a NetVLAD global feature. Considered that NetVLAD features did not take advantage of the spatial relationship between features, Jun Yu et al. [10] proposed Spatial Pyramid-Enhance NetVLAD(SPE-NetVLAD), which divides the image into multiple patches at different scales, and extracts NetVLAD features for each patch, finally, cascade all the NetVLAD feature. Noh et al. [11] considered the importance of features and added the Attention mechanism into the scene recognition task.

### 2.2 Point-Cloud based place recognition

With the gradual maturity of 3D scanning sensors (such as lidar, RGB-D sensors), 3D data is increasingly used in application scenarios such as autonomous driving, VR, and drones. The emergence of these 3D data usage has put forward the requirements for 3D data analysis algorithms. Some classic 3D point cloud feature extraction algorithms were proposed in the early days, such as Spin Images [12], Geometry Histogram [13], and recent work include Point Feature Histograms (FPH) [14], Fast Point Feature Histograms (FFPH) [15], Signature of Histograms of OrienTations (SHOT) [16]. These handcrafted feature descriptors are generally task-related and sensitive to noise and data incompleteness.

Considered that point cloud features should not be order-related. Charles R. Qi et al. [6] proposed the PointNet framework, the data manipulation in this framework is robust to point cloud order. Yangyan Li et al. [17] claimed that there is a potentially canonical order of points, based on which they proposed χ-Transformed operation to permute the point clouds.

The spatial relationship of the point cloud has an important role in features extraction. Roman Klokov et al. [18] proposed Deep Kd-Networks. They use the kd-tree of the point cloud to construct neural network, but this method is sensitive to the order of points. PointNet++ [19] proposed by Charles R. Qi et al. adds sampling and grouping layers to fuse spatial information. Yue Wang et al. [20] proposed DGCNN, which connects the K nearest neighbors of each point to form a graph for feature extraction. Compared to the connection relationship built in DGCNN, Hengshuang Zhao et al.[21] Proposed PointWeb, which use AFA Module to build full connections among neighbor points.

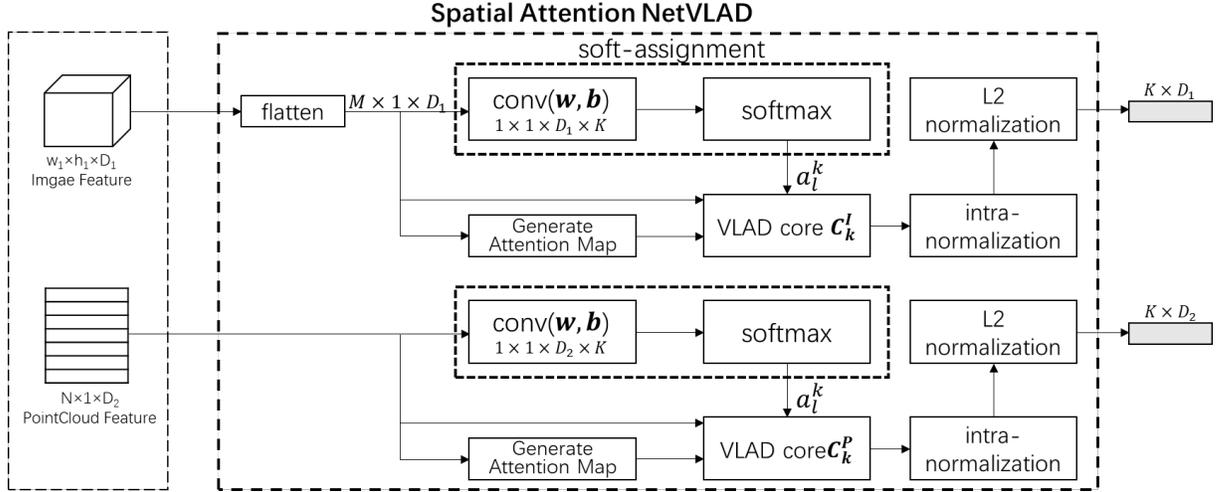

Figure 3.The Architecture of Spatial Attention NetVLAD (AttVLAD). Similar to NetVLAD, point cloud features can be input into AttVLAD directly, image feature should be flattened before input to AttVLAD,difierent to NetVLAD, when calculating the clustering center of each point or pixel, AttVLAD will learn an attention map and use it as the weight each point or pixel.

Recently, some researchers have used above method into the place recognition task, Mikaela Angelina Uy et al. [22] combined PointNet and NetVLAD, and proposed PointNetVLAD, which applied the idea of NetVLAD to point cloud feature extraction. Wenxiao Zhang et al. [23] considered that points have different importance in different regions, and added the Attention mechanism into PointNetVLAD. Local features play an important role in point cloud features, LPD-Net [7] adds handcrafted local feature and graph-based local feature aggregation module into the PointNetVLAD which significantly improve the performance.

## 2.3 Fusion features of point cloud and image

After the advent of RGB-D sensors (such as Microsoft Kinect), many computer vision related tasks began to use RGB-D sensor data. Tombari et al. [24] proposed CSHOT, which is based on SHOT [16] for feature matching. Eitel et al. [25] use RGB-D data for object recognition. Tang et al. [26] proposed MOANA for MOT in 3D. Ren et al. [27] use RGB-D data for pedestrian re-identification. Most of these works have one thing in common, they use depth information as a channel of the image and use image-based feature extraction methods to extract depth features. However, two adjacent pixels in the image may be far away in 3D space. As far as we know, in place recognition task, we may be the first attempt to fuse image features and point cloud features.

## 2. Method

### 3.1. Problem Definition

Similar to the image retrieval task, we have a scene database $D$, which contains $n$ scenes, each scene $M$ is consist of an image $I$ and a point cloud $P$. $\{t_{i1},...,t_{in}\}$ is the timestamp of each images, $\{t_{p1},...,t_{pn}\}$ is the timestamp of each point clouds, which ensure that:

$$t_{ik} - t_{pk} < t_{ij} - t_{pk}, k = \{1,...,n\}, \forall j \neq k \quad (1)$$

$\{p_1,...,p_n\}$ is the coordinate of each scene, and all scenes are evenly distributed in the real world. Each point cloud is downsampled to $N$ points with a downsampling filter $G(.)$ as PointNetVLAD [22] does.

**Definition.** Given a scene $q$, our goal is to find the most similar scene in database $M$.

### 3.2. Attention Based Collaboration

Image resolution is much higher than that of point cloud, but overexposure, darkness, weather and seasonal changes could cause information loss. Unlike changeable images, laser point clouds are robust to changes in illumination and texture, which can provide stable features for place recognition. In this section, we describe how to use the attention based collaboration method to fuse the stable point cloud feature and changeable image feature.

**Spatial Attention.** The local spatial attention module is used in both images and point clouds to select discriminative pixels and points, the details of the spatial attention module are shown in Fig.3.

As shown in the Fig.2, we use PointNet and LPD-Net (remove the NetVLAD layer) to extract the each point feature, denoted as Fp = $\{f_1,...,f_N\}$ ∈ $R^{N \times D1}$, and use ResNet50 (remove the final pooling layer) to extract image feature, denoted as Fi ∈ $R^{w \times h \times D2}$, where D1 and D2 are the feature dimension of point cloud and image. Our goal is to learn the spatial attention map A=$\{a_1,...,a_t\}$ ∈ $R^{t \times 1}$ of the image and the point cloud, where t = N for point cloud, and t = w×h for image, and add the attention map to the feature aggregation. similar to PCAN [23], NetVLAD is used to aggregate the local feature. NetVLAD learns $k$ cluster centers $\{c_1,...,c_k \mid c_k \in R^D\}$, and calculates the residual between feature $f$ and the cluster center $c$, each residual is then weighted by $a_l \times a^k_l$, where $a^k_l$ is also a learnable parameter, and $a_l$ is the attention weight of correspond pixel or point. The final spatial attention feature $V$ is expressed as:

$$V = [V1,...,Vk], \quad (2)$$

Where,

$$V_k = \sum_{l \in [1,N]} s_l a_l^k (f_l - c_k), \quad (3)$$

PCAN use a multi-scale SAG layer with the FP layer to extract the neighbor relationship and multi-scale features of point cloud, but it is computationally inefficient for the neighbor searching, we replace it with a $1 \times 1 \times D_1$ convolution layer. Similar to the point cloud, we use a $1 \times 1 \times D_2$ convolution layer for image attention map learning.

**Local Channel Attention.** As shown in the Fig.2, the feature from the point cloud branch is denoted as $G_P \in R^{Dp}$, and that from image branch is denoted as $G_I \in R^{Di}$, Before the fusion of the two, we further enhance each. The fully connected layer is used to learn the channel attention map of the $G_p$, $G_i$, and use this attention map to reweight the image and point cloud feature.

$$\boldsymbol{G'_p} = (G_p \times k_p) \times G_p, k_p \in R^{Dp \times Dp}, \quad (4)$$
$$\boldsymbol{G'_i} = (G_i \times k_i) \times G_i, k_i \in R^{Di \times Di}, \quad (5)$$

Where $kp$ and $ki$ are the learnable weights of the fully connected layer

**Global Channel Attention.** As we mentioned above, due to overexposure, dark night and et., sometimes most of the features of the image branch are unreliable, so does point cloud branch. Therefore, in the final fusion layer, it is meaningful to choose the reliable features from point cloud and image. Global Channel Attention is proposed to do this, similar to Local Channel Attention, fully connected layer is used to learn the global channel attention map, and use this global channel attention map to choose the reliable feature.

### 3.3. Day-Night Transfer for Color Normalization

Experiments show that the performance of the night sequence is significantly lower than other sequences, the change between day and night may be the biggest factor of failure retrieval. In order to bridge the gap, Style transfer tools, like CycleGan[28], pix2pix [29] could be used to do this job, we use CycleGan to learn the mapping function between day and night, which could transfer the color style of night to day.

## 3. Experiments

### 4.1. Datasets

To prepare the dataset, we use the generation method proposed in PointNetVLAD which uses Oxford RobotCar dataset. This dataset provides LIDAR point cloud, images from various directions, and trajectory. With the help of trajectory, LIDAR data can be merged into a complete scene map, each scene map is split into submaps in the forward direction. The ground points in the submaps have been removed, each submap is down sampled to 4096 points, normalized into [-1,1] and labeled with time stamp. We use this timestamp to find the image with the closest timestamp in the front camera, and form a pair with the point cloud submap. All pairs are split into training and test set using the same strategy proposed in PointNetVLAD. In the training set, adjacent submaps have and overlap of 10meters, and in the test set, there is no overlap. The final dataset has 21711 training pairs and 3030 test pairs.

### 4.2. Result

**Evaluation.** We extract the global feature of each place in the test set and form a place database. For each query, we search the database for the top-k similar place in Euclidean distance. If the coordinate of the return is within 25 meters of the query, we regard that is successful retrieval. The average recall of top-1% returns is used to evaluate the algorithm.

Table 1. Comparison results of direct concatenation

|  | Ave recall @ 1% |
|---|---|
| PN-VLAD | 80.31 |
| LPD-Net | 94.92 |
| RN-VLAD | 82.12 |
| PN-VLAD+RN-VLAD | 94.61 |
| LPD-Net + RN-VLAD | 97.70 |

Table 2. PIC-Net configuration

| PIC-01 | | PIC-02 | | PIC-03 | | PIC-04 | |
|---|---|---|---|---|---|---|---|
| Point Cloud + image/image in daytime | | | | | | | |
| PointNet/ LPD-Net | Renset50 | PointNet/ LPD-Net | Renset50 | PointNet/ LPD-Net | Renset50 | PointNet/ LPD-Net | Renset50 |
| NetVLAD | NetVLAD | AttVLAD | AttVLAD | VLAD | VLAD | AttVLAD | AttVLAD |
| Concat | | Concat | | Concat | | Concat | |
| - | | - | | Global Attention | | Global Attention | |
| L2-normalization | | | | | | | |
| Lazy Quadruplet Loss | | | | | | | |

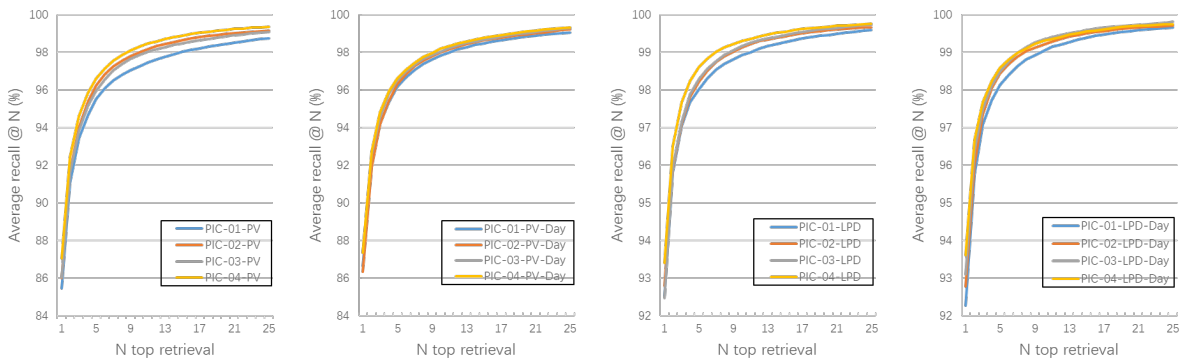

Figure 4. Average recall under different configuration

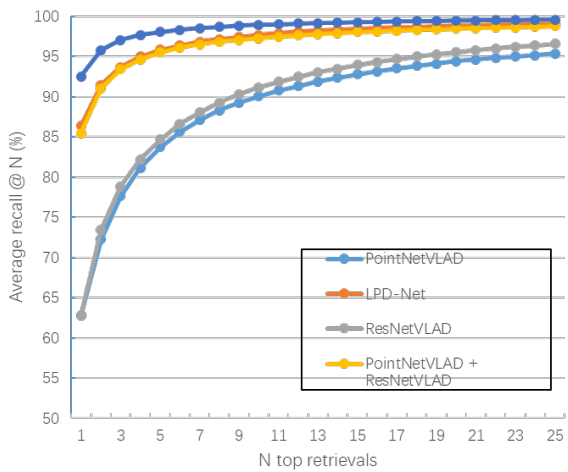

Firgre 5. Average recall of Direct Concatenation

Table 3. Comparison results of different configuration

| | recall @ 1% | recall @ 1% Day |
|---|---|---|
| PIC-01-PV | 94.61 | **95.32** |
| PIC-02-PV | 95.31 | **95.38** |
| PIC-03-PV | 95.13 | **95.81** |
| PIC-04-PV | 95.79 | **95.90** |
| PIC-01-LPD | 97.68 | **97.70** |
| PIC-02-LPD | 97.78 | **98.02** |
| PIC-03-LPD | 97.85 | **98.11** |
| PIC-04-LPD | 98.22 | **98.23** |

**Direct Concatenation.** We directly concatenate the feature of image and point cloud to form the global feature of the place, and compare this feature with ResNetVLAD(RN-VLAD), LPD-Net, and PointNetVLAD(PN-VLAD). The results are shown in Table 1, Fig.5. shows the average recall of top-25 retrievals. Which indicates that the direct concatenation could significantly improve the average recall. Through this experiment, we can see the complementarity between the image and point cloud, which could enhance each other.

**Optimal Configuration.** Table 2. is the configuration of PIC-Net, in each configuration, the backbone of point cloud feature could be either PointNet or LPD-Net (remove the VLAD layer). PIC-01 use NetVLAD for the feature aggregation, unlike PIC-01, PIC-02 replaces the NetVLAD layer with Spatial Attention NetVLAD, PIC-03 adds the global channel attention layer into the PIC-01, similar to PIC-03, PIC-04 adds the global channel layer into the PIC-02. In addition, for each configuration, we use both the original dataset and daytime-style dataset (night image is transfer to daytime-style) to train 2 models. All results are shown in Table 3, and Fig.4. is the average recall of top-25 retrievals. PIC-04-LPD-Day is the optimal configuration, which is 0.52% better than the direct concatenation.

Table 4. Comparison results of Branch Performance

|  | recall @ 1% | recall @ 1% of branch |
|---|---|---|
| PN-VLAD | 80.31 | **83.60** |
| PN-AttVLAD | 82.88 | **83.18** |
| LPD-Net | **94.92** | 93.89 |
| LPD-AttVLAD | 93.69 | **94.48** |
| RN-Vlad | 82.12 | **82.73** |
| RN-AttVLAD | 82.69 | **84.12** |

**Branch Performance.** As shown in Table 4, we observe that almost in all cases, branch of PIC-Net outperform original network, except LPD-Net. Which prove that the complementary information between image and point could propagate to each other through training.

### 4.3. Ablation Study

Table 5. Comparison results of Pixel Attention VLAD

|  | recall @ 1% |
|---|---|
| RN-VLAD | 82.12 |
| RN-AttVLAD | **82.69** |
| PN-VLAD+RN-VLAD | 94.61 |
| PN-VLAD+RN-AttVLAD | **95.10** |
| LPD-Net + RN-VLAD | 97.68 |
| LPD-Net + RN-AttVLAD | **98.20** |

**Spatial Attention.** Pixel Attention VLAD and Point Attention VLAD are two kind of Spatial Attention, proposed in this paper, we test the Pixel Attention VLAD in a variety of configurations. Results are in Table 5. we can see that, compared to RN-VLAD, RN-AttVLAD has an improvement of 0.57%. The gain of Pixel Attention VLAD still exists after fusion with point cloud, PN-VLAD + RN-AttVLAD is 0.49% better than PN-VLAD + RN-VLAD, and LPD-Net + RN-AttVLAD is 0.52% better than LPD-Net + RN-VLAD. These experiments indicate that the Pixel Attention VLAD could improve the recognition accuracy.

Table 6. Comparison results of Point Attention VLAD

|  | Ave recall @ 1% |
|---|---|
| PN-VLAD | 80.31 |
| PCAN | **83.81** |
| PN-AttVLAD | 82.88 |
| LPD-Net (no hard mining) | 93.42 |
| LPD-AttVLAD | **93.69** |
| PN-VLAD + RN-VLAD | 94.60 |
| PN-AttVLAD + RN-VLAD | **94.96** |
| LPD-Net + RN-VLAD | 97.68 |
| LPD-AttVLAD + RN-VLAD | **97.98** |

Similarly, we test the Point Attention VLAD in various configurations. From Table 6, we can see that PN-AttVLAD is 2.57% better than PN-VLAD. LPD-AttVLAD is 0.27% higher than LPD-Net. Compare to PCAN, PN-AttVLAD is 0.93% worse than PCAN, cause PN-AttVLAD only use a 1×1×D1 convolution layer to generate the attention map, which is much light than PCAN. Further results show that the gain of Point Attention VLAD still exists after fusion with image.

Table 7. Comparison results of Global Channel Attention

|  | recall @ 1% |
|---|---|
| PN-VLAD + RN-VLAD | 94.61 |
| PN-VLAD + RN-VLAD + GCA | **95.13** |
| PN-AttVLAD + RN-AttVLAD | 95.31 |
| PN-AttVLAD + RN-AttVLAD + GCA | **95.79** |
| LPD-Net + RN-VLAD | 97.68 |
| LPD-Net + RN-VLAD + GCA | **97.85** |
| LPD-AttVLAD + RN-AttVLAD | 97.78 |
| LPD-AttVLAD + RN-AttVLAD + GCA | **98.22** |

**Global Channel Attention(GCA).** The Global channel attention layer is used to finally weight the reliability between image and point cloud. As shown in Table 7, we test the Global channel attention in various configurations. Results shows that Global channel attention layer does work.

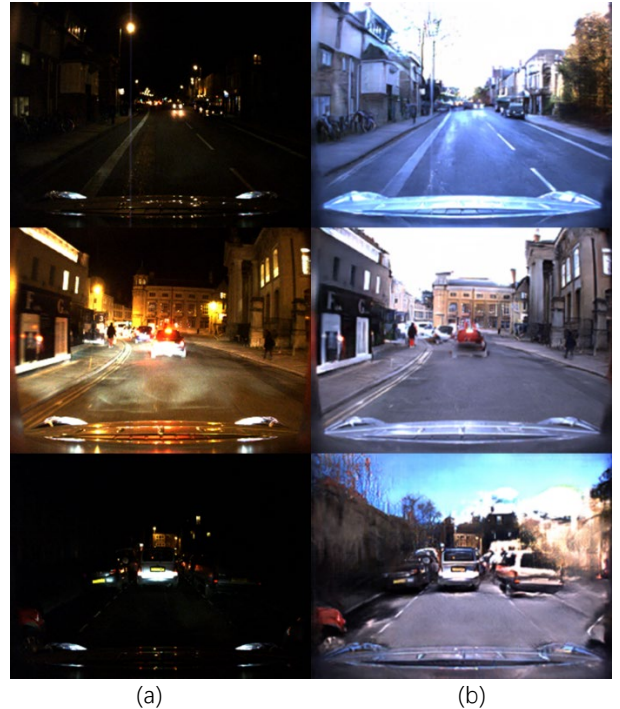

Figure 6. (a) are the real image collected in the night, (b) are the generated daytime image

Table 8. Comparison results of Day-Night transfer

| | recall @ 1% | recall @ 1% day |
|---|---|---|
| RN | 74.50 | 75.00 |
| RN-VLAD | 82.12 | 83.47 |
| RN-AttVLAD | 82.69 | 84.88 |
| LPD-AttVLAD + RNAttVLAD | 97.78 | 98.02 |
| PN-AttVLAD + RN-AttVLAD | 95.31 | 95.38 |

**Day-Night transfer.** Illumination change is the most important factor affecting the performance. Day-to-night transformation is used to normalize the color distribution under day and night conditions. As shown in the Fig.6, the degree of loss of information in night image is very different, and for images with little information loss, night images can be naturally converted to daytime style. For images where some information is lost, CycleGan can predict the lost information based on the retained information. For images where most of the information is lost, CycleGan can only convert the color back, but it cannot make up for the information that has lost. Table 8. gives the quantitative results that day-to-night transformation improves ResNet, RN-VLAD, and RN-AttVLAD by 0.5, 1.3, 2.18. Generally speaking, the normalization of color distribution could improve the final accuracy.

### 4.4. Results Visualization

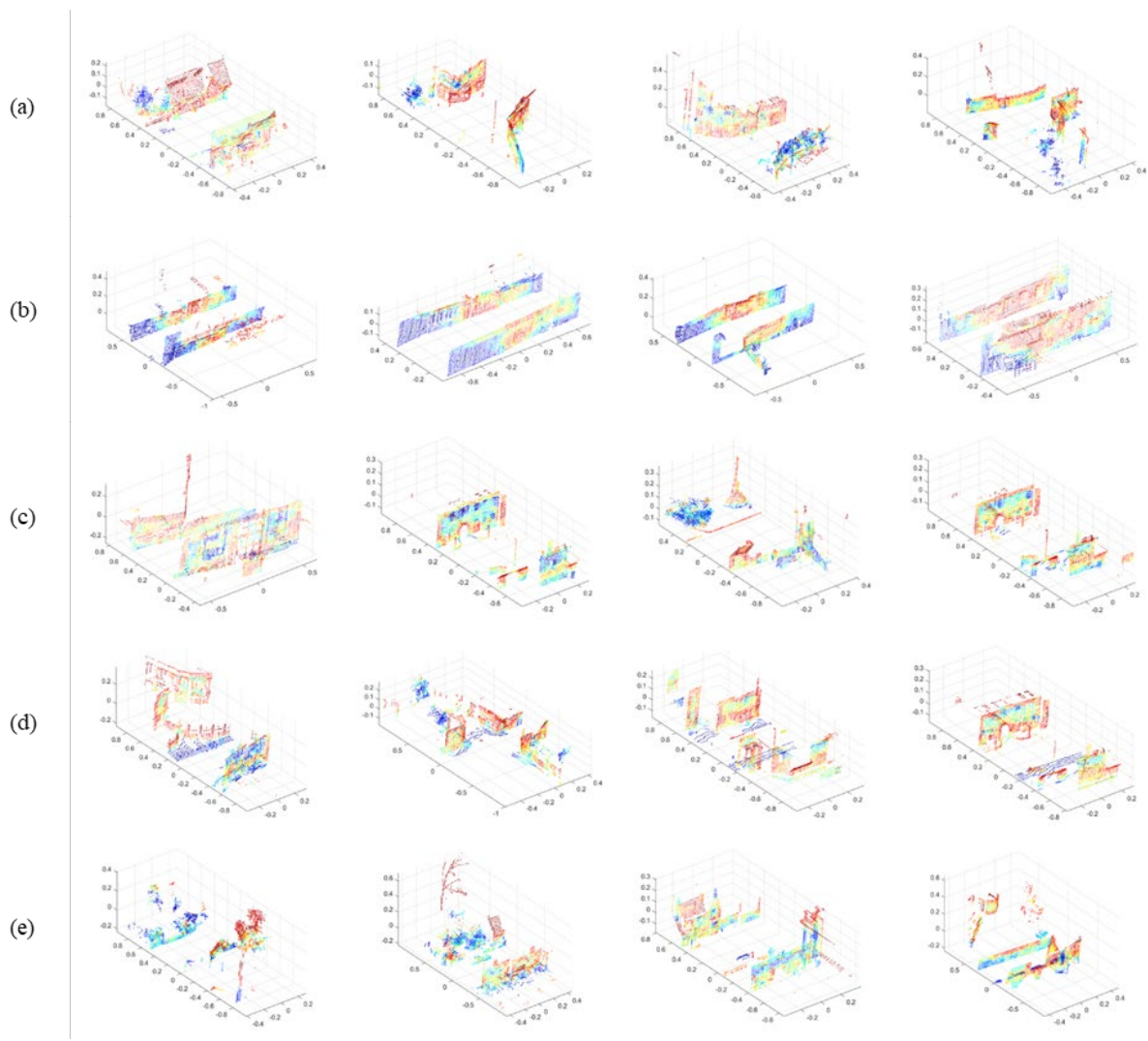

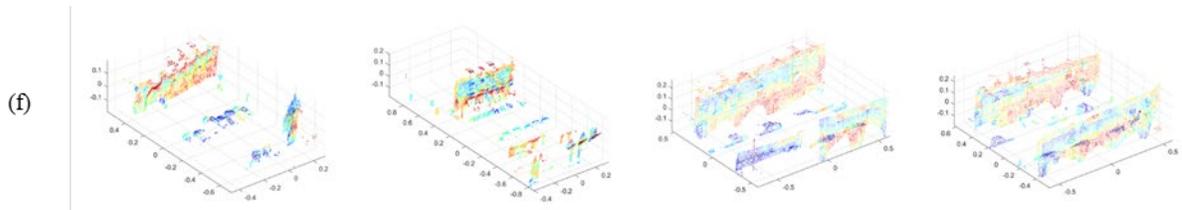

Figure 7. **Visualization of Point Attention.** (a) Leaves get lower weight and buildings get high weight. (b) center points get high weight and far point get low weight. (c) line structure get high weight and surface get lower weight. (d) ground is always ignored. (e) high point gets more attention. (f) cars are always ignored

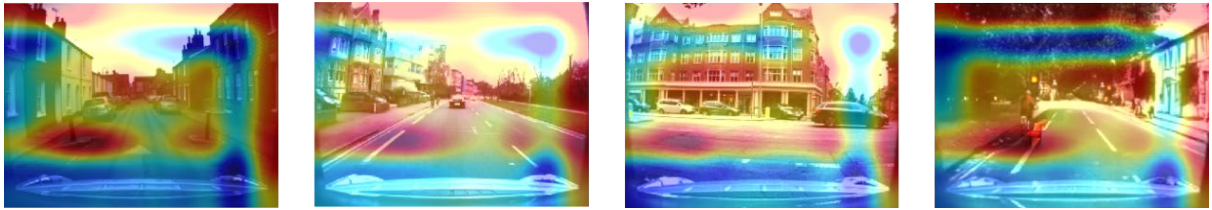

Figure 8. **Visualization of Pixel Attention.** Head of the car get little attention in all cases, sky get little attention in most cases, and the network pays more attention on the road.

**Point Attention.** We visualize the Spatial Attention of point cloud, which shows that the point cloud attention mechanism does provide clues to the feature extraction. As shown in Fig.7, Point Attention pays more attention on regular points, such as building, then clutter, such as leaves, and it pays more attention to close points than far points, furthermore, linear structures such as telegraph poles and signal towers get large weight than Planar structure, such as buildings. Ground is always ignored for its structureless. And high objects are often focused on for it unique.

**Pixel Attention**. Fig.8. is the visualization of Image Attention results, we can see that the bottom of all images (capture the front of car) are ignored by the Image Attention. In addition, the attention weight of the top position of most images is relatively small, because this position of images is always the sky. We observe that the road surface at the bottom left of the image always get the largest attention score and objects such as trees and land markings often appear there.

## 4. Conclusion

In this paper, we propose the Point cloud and Image Collaboration network, which uses both point cloud and image data for place recognition. Experimental results show that there is a strong complementarity between image and point cloud, and the Attention based fusion method could find discriminative points and pixels, and judge the reliability of the image and the point cloud. The final result can be further improved by night to day color normalization.